\documentclass[letterpaper]{article} % DO NOT CHANGE THIS
\usepackage{aaai24}  % DO NOT CHANGE THIS
\usepackage{times}  % DO NOT CHANGE THIS
\usepackage{helvet}  % DO NOT CHANGE THIS
\usepackage{courier}  % DO NOT CHANGE THIS
\usepackage[hyphens]{url}  % DO NOT CHANGE THIS
\usepackage{graphicx} % DO NOT CHANGE THIS
\usepackage{natbib}  % DO NOT CHANGE THIS AND DO NOT ADD ANY OPTIONS TO IT
\usepackage{caption} % DO NOT CHANGE THIS AND DO NOT ADD ANY OPTIONS TO IT
\usepackage{algorithm}
\usepackage{algorithmic}

\usepackage{todonotes}
\usepackage{multirow}
\usepackage{mathtools}
\usepackage{amsmath,amsfonts,amssymb}
\usepackage{xspace}
\usepackage{dsfont}
\usepackage{booktabs}
\usepackage{adjustbox}
\usepackage{tabularx}
\usepackage{enumitem}  % for leftmargin
\usepackage{pifont} % http://ctan.org/pkg/pifont
\newcommand{\xmark}{\ding{55}}%
\newcommand{\cmark}{\ding{51}}%	
\usepackage{colortbl}	
\usepackage{soul}
\usepackage{tcolorbox}

\usepackage[breaklinks=true,bookmarks=false]{hyperref}
\usepackage{ifsym}

\usepackage{wrapfig}

\newcommand{\code}[1]{\textcolor{darkgray}{\texttt{#1}}}

\newcommand{\answer}[1]{\textcolor{black}{\texttt{#1}}}

\newcommand{\saycanpay}{{SayCanPay}\xspace}

\DeclareMathOperator*{\argmax}{arg\,max}
\DeclareMathOperator*{\argmin}{arg\,min}

\newcommand{\psay}[1]{\ensuremath{p_{a_{#1}^i}^\text{say}}}
\newcommand{\pcan}[1]{\ensuremath{p_{a_{#1}^i}^\text{can}}}
\newcommand{\ppay}[1]{\ensuremath{p_{a_{#1}^i}^\text{pay}}}

\usepackage{newfloat}
\usepackage{listings}
\DeclareCaptionStyle{ruled}{labelfont=normalfont,labelsep=colon,strut=off} % DO NOT CHANGE THIS
\floatstyle{ruled}
\newfloat{listing}{tb}{lst}{}
\floatname{listing}{Listing}
\title{\saycanpay: Heuristic Planning with Large Language Models\\ using Learnable Domain Knowledge}

\author{
Rishi Hazra\textsuperscript{\rm 1} \quad
Pedro Zuidberg Dos Martires\textsuperscript{\rm 1} \quad
Luc De Raedt\textsuperscript{\rm 1,\rm 2}\\
}

\affiliations {
    \textsuperscript{\rm 1}Örebro University \quad
    \textsuperscript{\rm 2}KU Leuven\\
    \{rishi.hazra, pedro.zuidberg-dos-martires, luc.de-raedt\}@oru.se\\
    \Large\href{https://rishihazra.github.io/SayCanPay/}{\textcolor{magenta}{https://rishihazra.github.io/SayCanPay/}}
}

\begin{document}

\maketitle

\begin{abstract}
Large Language Models (LLMs) have demonstrated impressive planning abilities due to their vast ``world knowledge". Yet, obtaining plans that are both feasible (grounded in affordances) and cost-effective (in plan length), remains a challenge, despite recent progress. This contrasts with heuristic planning methods that employ domain knowledge (formalized in action models such as PDDL) and heuristic search to generate feasible, optimal plans. Inspired by this, we propose to combine the power of LLMs and heuristic planning by leveraging the world knowledge of LLMs and the principles of heuristic search. Our approach, \saycanpay, employs LLMs to generate actions (\emph{Say}) guided by learnable domain knowledge, that evaluates actions' feasibility (\emph{Can}) and long-term reward/payoff (\emph{Pay}), and heuristic search to select the best sequence of actions. Our contributions are (1) a novel framing of the LLM planning problem in the context of heuristic planning, (2) integrating grounding and cost-effective elements into the generated plans, and (3) using heuristic search over actions. Our extensive evaluations show that our model surpasses other LLM planning approaches. 
\end{abstract}

\section{Introduction}

% language models for reasoning and planning.
% classical planning vs language model planning (upsides and shortcomings)
% what we do
% how we do
% contributions

With the rise of Large Language Models (LLMs), there has been a growing interest in leveraging their generative capabilities for planning tasks~\cite{llms_zero_shot,llms_cant_plan,llm_guided_planning,llm_p}. These models have the ability to generate long-horizon plans, capitalizing on their extensive ``world knowledge" gained from training on vast amounts of data (e.g. eggs are typically stored in the refrigerator, and placing an apple in the fridge will cool it). Such expansive knowledge can be exploited to plan in an open-world context~\cite{ding2023integrating}. Moreover, planning in the natural language space offers significant flexibility especially, with the advent of multimodal foundation models~\cite{speechLMs,vlm_survey,rt2-vla}. Such models have made it easier to represent various modalities such as vision, speech, and even actions in the form of natural language, thus bypassing the need to have domain-specific knowledge (e.g. PDDL) that traditional planning approaches require. However, LLM-based planning often faces challenges, particularly in generating feasible plans. It can fail to model action affordances (or pre-conditions)\footnote{In robotics, affordances refer to possible actions that \emph{can} be executed, which is conceptually similar to inferring preconditions in planning -- what actions are feasible in a certain situation.} due to difficulty in modeling the state of the world (e.g. \emph{grab milk from the fridge} even if the door is closed) or having a pretrained world model that is not aligned with the current environment (e.g. \emph{using a controller to regulate the heater} where only a knob exists), leading to infeasible plans. Moreover, such models focus greedily on the next actionable step without considering its relevance to the ultimate goal, resulting in longer, cost-inefficient plans~\cite{valmeekam2023planning}. Recent works like SayCan~\cite{saycan} have sought to address the affordance problem by using pretrained skills to evaluate the action's executability -- \emph{Can the action be executed in the current state?} However, the plan cost remains a concern.

%%%%%%%%%%%%%%%%%%%%%%%%%%%%%%%%%%%%%%%%%%%%%%%%%%
\begin{figure}[ht]
    \centering
    \includegraphics[width=\linewidth]{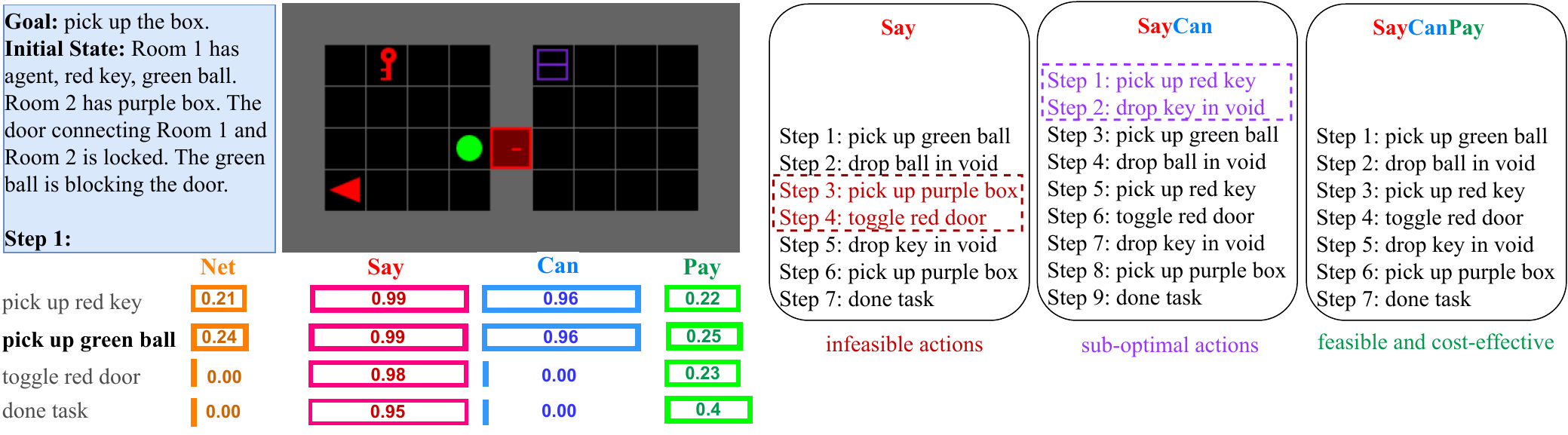}
    \caption{Figure illustrates how SayCanPay scores the next action in BabyAI environment~\cite{babyai_iclr19}. Given inputs: goal $g$ and initial observation $o_0$, the Say model generates candidate actions with associated probabilities. These are then scored for feasibility by the Can model and for payoff by the Pay model. Here, the Can model deems both \emph{pick up red key} and \emph{pick up green ball} equally probable (i.e. both preconditions are satisfied). However, the Pay model ensures a better payoff for \emph{pick up green ball}. We compare plans generated by Say, SayCan, and SayCanPay scoring. Say scoring can lead to infeasible plans and SayCan to feasible but longer plans. The displayed grid is purely illustrative, with no visual inputs used.}
    \label{fig:saycanpay_teaser}
\end{figure}
%%%%%%%%%%%%%%%%%%%%%%%%%%%%%%%%%%%%%%%%%%%%%%%%%%

In contrast, traditional planning provides an established approach to developing a sequence of actions to transition from an initial state to a goal state. It uses a domain file (with action models defined in PDDL specifying pre- and post-conditions) and heuristic search planners like Fast Downward~\cite{fastdownward_planning} to ensure feasibility through grounding in preconditions, and generating cost-effective plans by employing search trees to select the best (or shortest) sequence of actions. However, obtaining a domain file for complex real-world environments is difficult, and its use restricts planning to a closed-world setting. These methods also struggle to handle partial observations, although approximate planning~\cite{kaelbling1998planning} can alleviate it.

Integrating LLMs with classical planning offers a promising research path, merging the generative abilities and (open) world knowledge of LLMs with the methodological rigor of planning algorithms. To this end, we extend the following contributions. \textbf{(1)} We propose to frame language model planning in the context of heuristic planning, which to our knowledge, is the first of its kind (\S~\ref{section:planning_with_language_models}). \textbf{(2)} We incorporate feasibility and cost-effective elements into the generated plans using a joint scoring named \textbf{\saycanpay}. As shown in Figure~\ref{fig:saycanpay_teaser}, it guides the planning through three key steps: (i) Say: Given a goal and an initial observation, the LLM generates \emph{likely} candidate actions at each step; (ii) Can: An affordance model scores these actions' {\em feasibility}, mirroring the evaluation of preconditions; (iii) Pay: Another model scores the actions according to their {\em estimated payoff}, akin to heuristic estimators (\S~\ref{section:saycanpay_inference}). The Can and Pay models undergo domain-specific training to align the plans with the current environment (\S~\ref{section:saycanpay_training}). \textbf{(3)} Using this combined score as a heuristic, we search for the most feasible and cost-effective plan (\S~\ref{subsection:beam_actions}). We demonstrate how our proposed joint scoring and heuristic search improve over the current LLM planning frameworks (\S~\ref{subsection:results}).
%%%%%%%%%%%%%%%%%%%%%%%%%%%%%%%%%%%%%%%%%%%%%%%%%%%%%%%%%%%%%%%%
\section{Related Work on Planning with LLMs}
\label{section:related_work}

%%%%%%%%%%%%%%%%%%%%%%%%%%%%%%%%%%%%%%%%%%%%%%%%%%%%%%%%%%%%%%%%%%%%%%%%%%

\begin{table}[ht]
\small
\centering
\begin{tabular}{lcccccc}
% \toprule
\textbf{Model} & \textbf{I/O} & \multicolumn{1}{c}{\textbf{Planner}} & \multicolumn{2}{c}{\textbf{Domain Knowledge}} & \textbf{Search} & \textbf{Planning} \\
 & &  & Affordances & Heuristics &  &  \\
 \hline
HSP~\cite{heuristic-search} & Symbolic & Symbolic & \cmark & \cmark & Heuristic & Offline \\
LLM+P~\cite{llm_p} & Hybrid & Symbolic & \cmark & \cmark & Heuristic & Offline \\
\hline
Planning LM~\cite{llms_zero_shot} & NL & LLM & \xmark & \xmark & Greedy$^{\ast}$ & Offline \\
SayCan~\cite{saycan} & NL & LLM & \cmark & \xmark & Greedy$^{\ast}$ & Online\\
Grounded Decoding~\cite{grounded_decoding} & NL & LLM & \cmark & \xmark & Greedy$^{\ast}$ & Online \\
Text2Motion~\cite{text2motion} & NL & LLM & \cmark & \xmark & Greedy$^{\ast}$ & Online\\
% \rowcolor{lightgray}
ProgPrompt~\cite{progprompt} & Symbolic & LLM & \cmark & \xmark & Greedy$^{\ast}$ & Offline\\
% \rowcolor{lightgray}
Plansformer~\cite{plansformer} & Symbolic & LLM & \cmark & \xmark & Greedy$^{\ast}$ & Offline \\
% \rowcolor{gray}
\hline
\textbf{SayCanPay} (Beam-Action) & NL & LLM & \cmark & \cmark & Heuristic & Offline\\
\hline
\end{tabular}
\caption{Table contrasts \saycanpay with existing works. I/O: input (goal/task, observation/state) / output (actions), NL: natural language. Here, Greedy$^{\ast}$ suggests the algorithm greedily selects actions while (possibly) searching over tokens.}
\label{tab:related-works-comparison}
% \vspace{-9pt}
\end{table}
% \rowcolor{lightgray}
%%%%%%%%%%%%%%%%%%%%%%%%%%%%%%%%%%%%%%%%%%%%%%%%%%%%%%%%%%%%%%%%%%%%%%%%%%%%

% \paragraph{Planning with Large Language Models.}
% lm generating plan
% pddl-based planning with translation
% grounded planning

Table~\ref{tab:related-works-comparison} categorizes LLM planning works into two broad categories based on whether the inputs (goals, states) and output actions (I/O) are natural language (NL) or symbolic (PDDL, scripting language). The approaches in the first category~\cite{llms_zero_shot,llms_cant_plan} often fail to model action affordances and the state of the world, leading to the generation of infeasible plans~\cite{llms_cant_plan}. To improve the groundedness, recent works have explored planning guided by learnable domain-specific models that score the actions' \textit{feasibility} akin to preconditions~\cite{grounded_decoding,text2motion}. Notably, SayCan \cite{saycan} uses pretrained low-level skills to ground the LM-generated actions. Others have used online planning with environmental and human feedback~\cite{innermonologue}. A limitation of such models, however, is their short-sighted nature, as they focus greedily on the next feasible action without considering its long-term relevance to the goal. Moreover, the plans are generated in an online fashion, interleaving action generation and execution, thus simplifying state tracking. In contrast, \saycanpay performs offline planning (i.e. complete plan generation while maintaining an internal world state) with both precondition and heuristic estimators, improving plan feasibility and cost-efficiency.

Another line of work employs LLMs to create offline symbolic plans, leveraging LLMs' training on open-source codebases, where actions appear as function calls~\cite{progprompt,codeaspolicies}. The feasibility of plans is ensured through assertion checks (\emph{assert $\langle$ preconditions $\rangle$}), that may trigger recovery actions. However, it relies solely on the LLM's domain knowledge which is limited to its training data and may not be aligned with the agent's current environment (e.g. espresso machine operations vary widely). Conversely, \saycanpay uses additional models trained with domain-specific knowledge collected from the current environment. There are also efforts to fine-tune LLMs like Code-T5~\cite{codet5} to generate plans in PDDL~\cite{plansformer}. This requires a significant amount of training data (given LLMs' minimal PDDL exposure) which is not entirely justified by their performance. 

Yet another exciting line of work explores hybrid I/O systems like LLM+P~\cite{llm_p} wherein, given a PDDL domain file (with a predefined action model), the LLM maps the NL inputs (task description, input observation) to a PDDL problem file. A symbolic planner then generates the plan. However, its effectiveness is limited by the closed-world constraint of the domain file, the necessity for fully observable states, and the LLM's restricted capability in translating NL to PDDL~\cite{language_to_goals}.
%%%%%%%%%%%%%%%%%%%%%%%%%%%%%%%%%%%%%%%%%%%

\section{Preliminaries}
\label{section:preliminaries}

\subsubsection{Planning Framework.}\label{subsubsection:planning_framework} We formulate our planning problem, based on approximate planning~\cite{golowich2022planning}, as a finite-horizon Partially Observable Markov Decision Process (POMDP) given by the tuple $\langle \mathcal{S}, \mathcal{S}_G, b_0, \mathcal{A}, \mathcal{O}, R, \mathbb{T}\rangle$. Here, $\mathcal{S}$ is state space, $\mathcal{S}_G \subseteq \mathcal{S}$ is a set of goal states, $b_0$ is the initial belief state, $\mathcal{A}$ is the set of actions, $\mathcal{O}$ is a set of observations retrieved from states via an observation function $\mathrm{O}$, $R: \mathcal{O} \rightarrow \mathbb{R}$ is a known reward function, $\mathbb{T}: \mathcal{S} \times \mathcal{A} \rightarrow \Delta^{\mathcal{S}}$ is a known stochastic transition function and $\Delta^{\mathcal{S}}$ is a distribution over states. Belief states represent the agent's knowledge of the environment at any point, given as $b \in \Delta^{\mathcal{S}}$. Additionally, let $\mathcal{H}_t\coloneqq(\mathcal{A} \times \mathcal{O})_{t-1}$ denote the set of \textit{histories} at step $t$, namely the set of action/observation sequences $(o_0, a_1, o_1, \dots, a_{t-1}, o_{t-1})$ or $(a_{1:t-1}, o_{0:t-1})$ the agent has access to before selecting action $a_{t}$. It is assumed that the goal states are fully observable. 

Unlike MDPs, the optimal policy in a POMDP typically takes actions depending on not just the most recent observation but the entire history. The objective of the planning algorithm is to find the optimal sequence of actions $a_{1:T}$ (i.e. an optimal plan) from an initial belief state $b_0$ to a given goal state $g \in \mathcal{S}_G$. Here, $T$ is the length of the horizon.
% focusing on parts of the search space that are more likely to lead to a goal.

\subsubsection{Heuristic Search Planning.}\label{subsubsection:planning_as_heuristic_search} In real-world scenarios where the state space can be exponentially large to explore exhaustively, heuristic search planning (HSP) becomes useful~\cite{heuristic-search}. Essentially, it uses heuristic functions $f_{\text{heur}}: \mathcal{H}_t \times \mathcal{S}_G \rightarrow \mathbb{R}$ to guide the search process in the planning problem, by computing a cost estimate from a given history of actions and observations. An example is the Best-First Search algorithms that select the most promising (next) action(s) using a linear combination of previously accumulated \textbf{cost} $f_{\text{acc}}$ for history $h_{t-1}$, and the estimated \textbf{cost} $f_{\text{heur}}$ from updated history $h_{t} = (h_{t-1}, a_{t})$ and goal $g$.

\begin{equation}
    f(h_{t}) = z_1 \cdot f_{\text{acc}}(h_{t-1}) + z_2 \cdot f_{\text{heur}}(h_{t}, g)
\label{eq:heuristic_search}
\end{equation}

 Here $z_1$, $z_2$ $\in \{0,1\}$. The next action $a_t = \argmin_{h_t} f(h_t)$. Special cases are the $A^{\ast}$ algorithm algorithm ($z_1=1$ and $z_2=1$) and Greedy Best-First Search ($z_1=0$ and $z_2=1$).

%%%%%%%%%%%%%%%%%%%%%%%%%%%%%%%%%%%%%%%%%%%
\begin{figure*}[t]
    \centering
    \includegraphics[width=0.9\linewidth]{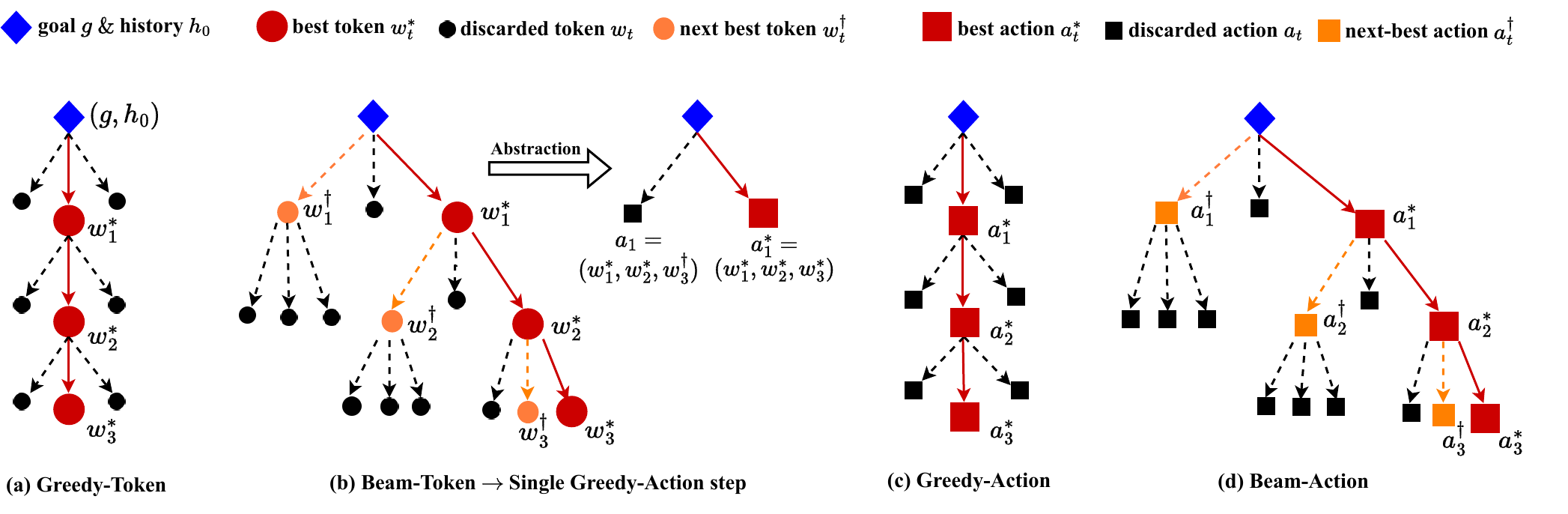}
    \caption{The figure outlines decoding strategies -- Greedy-Token, Greedy-Action, and Beam-Action. Greedy-Token greedily selects the next best token by its probability. Greedy-Action (which is a beam search over tokens) greedily selects the next best action based on a specific decoding score. Beam-Action uses a beam search over actions, maintaining $k$ beams and selecting the best sequence as the plan. Here, nodes represent either tokens $w_t$ or actions $a_t$. The best plan is given by $(a_1^\ast, a_2^\ast, a_3^\ast)$ and represented in red. The second-best node is in orange, discarded ones in black. Here, for Beam-Action, $m=3$ and $k=2$.}
    \label{fig:decoding_strategies}
\end{figure*}
%%%%%%%%%%%%%%%%%%%%%%%%%%%%%%%%%%%%%%%

\section{Language Model Planning Framework}
\label{section:planning_with_language_models} 
We keep the same POMDP formulation while updating our interpretations of the tuple. Previous works have shown that language models (LMs) trained on extensive data would internalize rich world knowledge that can be queried for downstream tasks like planning~\cite{hao2023reasoning}. This is akin to an internal transition function $\mathbb{T}^{int}$. Similarly, LMs also maintain and update an internal belief state $b_t^{int}$ over tokens (or actions). An observation function maps states to NL observations, $\mathrm{O}: \mathcal{S} \rightarrow \mathcal{O}$. The updated POMDP is now given as $\langle \mathcal{S}, \mathcal{S}_G, b_0^{int}, \mathcal{A}, \mathcal{O}, R, \mathbb{T}^{int} \rangle$. In our offline planning experiments, we assume the following:~(i)~$\mathcal{O}=\{o_0\}$ inducing belief state $b_0^{int} = \mathds{1}_{s_0}$, while $o_t = \emptyset \; \forall \; t > 0$, due to lack of environmental feedback;~(ii)~sparse rewards $=1$ for plan success, else $0$. While our LM does not utilize the reward function, one could use it for alignment~\cite{RLHF}.

\noindent \textbf{Problem Statement}: Given a NL goal $g$, history $h_0=(o_0)$, and a LM generating actions $a_t$ with probability $p(a_t | h_{t-1}, g)$, generate the most likely plan ($a_{1:{T}}$) to go from $b_0^{int}$ to $g$, i.e., $\arg \max_{{a_{1:{T}}}} P(a_{1:{T}}|h_0, g)$. 

\noindent We aim to maximize the plan's probability, reframing LM planning as a classical search problem, where we repeatedly expand the current plan $a_{1:{t-1}}$ by adding action $a_{t}$. Rewriting the probability $P(a_{1:T} | h_0, g)$ recursively as:

\begin{equation*}
\begin{split}
    &= P(a_{1:t-1}, a_t, a_{t+1:T} | h_0, g) \\
    &= p(a_{1:t-1} | h_0, g) p(a_t | h_0, a_{1:{t-1}}, g) p(a_{t+1:T} | h_0, a_{1:t}, g) \\
    &= p(a_{1:t-1} | h_0, g) \cdot p(a_t | h_{t-1}, g) \cdot p(a_{t+1:T} | h_t, g)
\end{split}    
\end{equation*}

To align with Eq~\ref{eq:heuristic_search} of the planning problem, we take $\log$ on both sides and \textbf{maximize} rather than minimize. We get accumulated \textbf{value} $f_{\text{acc}}(h_{t-1}) = \log p(a_{1:t-1} | h_0, g)$, heuristic \textbf{payoff} $f_{\text{heur}}(h_{t}, g) =  p(a_{t+1:T} | h_t, g)$, and $f(h_{t}) = \log P(a_{1:T} | h_0, g)$. Rewriting the above equation:

\begin{equation}
    f(h_{t}) = f_{\text{acc}}(h_{t-1}) + \log \big(p(a_t | h_{t-1}, g) \cdot f_{\text{heur}}(h_{t}, g)\big)
\end{equation}

The additional $p(a_t | h_{t-1}, g)$ reflects that, unlike classical planning which evaluates only feasible actions based on preconditions, LMs assign probabilities to each action. Here, next action $a_t = \argmax_{h_t} f(h_t)$.

Technically, the LM generates actions wherein each action is a sequence of tokens until the end-of-sequence token, $\langle \text{EOS} \rangle$. For each action step $a = (w_1, \dots, w_n)$ composed of tokens $w_i$, the LM computes the action probability as $p(a) = p(w_1)\prod_{i=2}^{n} p(w_i | w_{1:i-1})$. Planning LM~\cite{llms_zero_shot} proposed a greedy decoding strategy wherein the LM greedily picks the next token, henceforth referred to as \textbf{Greedy-Token} baseline (Figure~\ref{fig:decoding_strategies} Left). The generated action is then appended to the history $h_{t}$= $(h_{t-1}, a_{t})$, and the generation process repeats until a ``\emph{done task}" action is generated. Subsequent works~\cite{text2motion} have investigated beam search over tokens. However, we are mainly interested in searching on the level of actions and not tokens.
%%%%%%%%%%%%%%%%%%%%%%%%%%%%%%%%%%%%%%%%%%%%%%%%%%%%%%%

%%%%%%%%%%%%%%%%%%%%%%%%%%%%%%%%%%%%%%%%%%%%%%%%%%%%%%%%%%%%%%%%%%
\section{\saycanpay Inference}
\label{section:saycanpay_inference}

The core concept of \saycanpay is to guide LMs in generating feasible and cost-effective plans. The process unfolds in three key steps: (1) \textbf{Say}: At each step $t$, the LM generates the \emph{top-m} candidate actions with associated probabilities $\{p(a_t^i | h_{t-1}, g)\}_{i=1}^{m}$. This generation employs a beam search over tokens. (2) \textbf{Can}: Next, a trained domain-specific model weighs these candidate actions on their feasibility, mirroring precondition evaluation. (3) \textbf{Pay}: Finally, a trained domain-specific estimator weighs the candidate actions according to their estimated payoff. The probabilities from these three components are then combined to select the next action. An overview of \saycanpay is provided in Figure~\ref{fig:saycanpay_teaser}.

In what follows, we instantiate the LM planning problem with two decoding strategies (or search algorithms that select the next action(s)): \textbf{Greedy Action} (\S~\ref{subsection:greedy_actions}) and \textbf{Beam Action} (\S~\ref{subsection:beam_actions}). Each strategy is explored using three distinct decoding scores (i.e. score used by the search algorithm to select the next action) -- Say, SayCan, SayCanPay. We then elaborate on the training of Can and Pay models (\S~\ref{section:saycanpay_training}).

%============================================================================
\subsection{Greedy-Action}
\label{subsection:greedy_actions}

In this decoding strategy, we maintain a single action sequence and at each step, greedily choose the next \emph{best} action based on a specific decoding score. This is akin to performing Greedy Best-First Search with $z_1=0$ and $z_2=1$. The decoding score for each candidate action $a_t^i$ is given as:
\begin{equation*}
    f(h_{t}^i) = \log \big(p(a_t^i | h_{t-1}, g) \cdot f_{\text{heur}}(h_{t}^i, g)\big)
\end{equation*}

Here, the best action $a_t^{\ast} = \argmax_{h_t^i} f(h_{t}^i)$, where $h_{t}^i = (h_{t-1}, a_t^i)$ denotes the current history with $i^{th}$ candidate action. As shown in Figure~\ref{fig:decoding_strategies}, this approach can be viewed as being ``greedy" with respect to actions while using ``beams" over the tokens. Now, we explore three variations of the strategy based on how the decoding score is computed. 

\begin{itemize}
    \item \textbf{Say}: In this decoding score, we set the estimated payoff $f_{\text{heur}}(h_{t}^i, g)=1 \; \forall \; i \in \{1, \dots, m\}$. Hence, the action is selected solely based on the LM generation probability, without considering feasibility or payoff.
    
            \begin{equation}
                f(h_{t}^i) = \log\big(\underbrace{p(a_t^i | h_{t-1}, g)}_{\eqqcolon \psay{t}}\big)
            \end{equation}

    \item \textbf{SayCan}: Here, the action feasibility is also considered. Let, $\sigma_t = (a_t, pre(a_t))$ where $pre(a_t)$ denotes the preconditions of $a_t$. The ``can" probability\footnote{The goal $g$ is used to evaluate the preconditions of ``done task".}, is denoted by $p(pre(a_t) | h_{t-1}, g)$. Again, $f_{\text{heur}}(h_{t}^i, g)=1 \; \forall \; i$.
    
        \begin{equation}
        \begin{split}
            f(h_{t}^i) &= \log\big(p(\sigma_t^i | h_{t-1}, g)\big) \\
            & = \log\big(\underbrace{p(a_t^i | h_{t-1}, g)}_{\eqqcolon \psay{t}} \cdot \underbrace{p(pre(a_t^i) | h_{t-1}, g)}_{\eqqcolon \pcan{t}}\big)
        \end{split}
        \end{equation}

    \item \textbf{SayCanPay}: This decoding score accounts for the estimated payoff in addition to the abovementioned scores. Hence, the best action is selected based on a combined score of Say, Can, and Pay scores.
    
        \begin{equation}
            \log\big(\underbrace{p(a_t^i | h_{t-1}, g)}_{\eqqcolon \psay{t}} \cdot \underbrace{p(pre(a_t^i) | h_{t-1}, g)}_{\eqqcolon \pcan{t}} \cdot \underbrace{f_{\text{heur}}(h_{t}^i, g)}_{\eqqcolon \ppay{t}}\big)
        \end{equation}
\end{itemize}

%============================================================================
%\footnote{A balance is sought between computational resources and the ability to explore a wide range of actions, so only $k$-beams are maintained, rather than considering all possible candidate actions.}
\subsection{Beam-Action}
\label{subsection:beam_actions}
In heuristic planning, multiple potential plans (i.e. action sequences) are simultaneously maintained and iteratively expanded until the goal is achieved. To simulate this behavior, we propose to manage $k$ action sequences. It works as follows -- each sequence is expanded with $m$ candidate actions (where $m \geq k$) from the LM, resulting in a total of $k \times m$ sequences. Then, top-$k$ sequences are retained using a specific decoding score accumulated over the sequence, as shown below. Once all $k$-beams have terminated, we select the sequence with the highest (length-normalized)\footnote{Since different beams can have different sequence lengths.} accumulated score. To avoid repetition, we only show the SayCanPay version. The rest can be similarly formulated.
\begin{equation*}
    \text{top-}k \bigg[\frac{1}{|h_{t}^{ij}|}\bigg(f_{\text{acc}}({h_{t-1}^i}) + \log p(\sigma_t^j | h_{t-1}^i, g) \cdot f_{\text{heur}}(h_{t}^{ij}, g)\bigg)\bigg]
\end{equation*}
Here, $i \in \{1, \dots, k\}$, $j \in \{1, \dots, m\}$, $k \leq m$. The updated history $h_t^{ij} = (h_{t-1}^{i}, a_t^{j})$ is obtained by adding the action $a_t^j$ to the $i^{th}$ beam history $h_{t-1}^{i}$. The outcome becomes the value for $f_{\text{acc}}(h_t)$ for the next iteration. Note, that setting $k=1$ results in Greedy-Action decoding.
% Simply, we start with histories $\{h_t^i\}_{i=1:k}$ and are left with updated histories $\{h_t^l\}_{l=1:k}$ after a single step of Beam-Action decoding.

Our proposed decoding has similarities with Tree-of-Thoughts inference~\cite{tree-of-thoughts} which also maintains multiple reasoning paths to decide the next step. However, our method is specifically tailored for planning problems. It uses search and evaluation techniques akin to planning methods, making it more suited for such challenges. Now, we discuss the training details of the Can and Pay models.
%============================================================================
\section{Learning the Can and Pay Models}
\label{section:saycanpay_training}

To train our domain-specific Can and Pay models, we collect $N$-expert trajectories $\mathcal{E} = \{\tau\}_{n=1}^{N}$ for each environment using an oracle planner, where $\tau_i = (o_0, g, a_1, a_2, \dots, a_T, r)$. Note, $r=1$ for all expert trajectories.

\subsection{Can Model}
\label{subsection:can_model}
We model it as a classification problem, where the positive action (i.e., the action whose preconditions are satisfied) is assigned the highest probability from a set of one positive and a few negative actions. Specifically, we sample a batch of actions $[h_{t-1}, g, a_t, a_{\bar{t} \neq t}, \tilde{a}]^{1:B}$ from expert trajectories $\mathcal{E}$. We then train a model $\mathcal{M}^{\text{can}}$ with the aim of minimizing the InfoNCE loss~\cite{infoNCE_loss}:
\begin{align*}
    -\frac{1}{B} \sum_{i=1}^{B} \log
    \frac{
            \mathcal{M}^{\text{can}}(h_{t-1}^i, g^i, a_t^i)
        }
        {
            \sum_{a\in \{a_t^i, a_{\bar{t} \neq t}^i, \tilde{a}^i \}}
            \mathcal{M}^{\text{can}}(h_{t-1}^i, g^i, a)
        }
\end{align*}

Here, $B$ is the batch size, $a_t$ is the positive action from trajectory $\tau_i$ executed in the context of history $h_{t-1}$ with goal $g$, $a_{\bar{t} \neq t}$ is a negative action sampled from the same trajectory $\tau_i$, but at a different time-step $\bar{t}$, and $\tilde{a}$ is a negative action sampled from a different trajectory $\tau_{j \neq i}$ with a different initial observation $o_0$ and goal $g$. $\mathcal{M}^{\text{can}}$ consists of an uncased Bert model~\cite{bert2019} with a probe layer and is trained end-to-end to correctly identify the positive action. The input to $\mathcal{M}^{\text{can}}$ is of the format `$\langle \mathrm{Goal} \rangle \{g\} \; \langle \mathrm{History} \rangle \{h_{t-1}\} \; \langle \mathrm{NXT} \rangle \{a_t\}$'. Here, `$\langle \ast \rangle$' serves as special tokens. The output is the Can probability $p_{a_t}^\text{can} \coloneqq \mathcal{M}^{can}(h_{t-1}, g, a_t)$. The model is trained across multiple batches for F1-score convergence on the validation set. Our approach is different from SayCan~\cite{saycan} which trains multiple affordance functions (corresponding to different skills), through temporal-difference-based reinforcement learning to predict the likelihood of a particular skill succeeding (i.e., executing) in the current state. Here, we show two training I/O examples, one with positive action and another one with negative action.\\

\noindent \textbf{Input} \code{$\langle$Goal$\rangle$} pick up the purple box. \code{$\langle$Initial State$\rangle$} Room 1 has yellow key, agent. Room 2 has purple box. The door connecting Room 1 and Room 2 is locked. \code{$\langle$Step 1$\rangle$} pick up yellow key. \code{$\langle$NXT$\rangle$} toggle yellow door.\\
\textbf{Output} \answer{1.0}  \quad\quad// \emph{feasible}

\noindent \textbf{Input} \code{$\langle$Goal$\rangle$} pick up the purple box. \code{$\langle$Initial State$\rangle$} Room 1 has yellow key, agent. Room 2 has purple box. The door connecting Room 1 and Room 2 is locked. \code{$\langle$Step 1$\rangle$} pick up yellow key. \code{$\langle$NXT$\rangle$} pick up purple box.\\
\textbf{Output} \answer{0.0}  \quad\quad// \emph{infeasible}

%============================================================================

%%%%%%%%%%%%%%%%%%%%%%%%%%%%%%%%%%%%%
\begin{table}[t]
\small
  \centering
    \begin{tabularx}{\textwidth}{p{3cm}p{2.0cm}p{7.5cm}>{\centering\arraybackslash}p{1.7cm}>{\centering\arraybackslash}p{0.5cm}}
\toprule
    \textbf{Environment} & \textbf{Example Goal} & \textbf{Example Initial Observation} &  \textbf{Plan Length} & $\mathbf{|\mathcal{A}|}$ \\
\midrule
Ravens (Tower~of~Hanoi~seq) & Move the gray disk in rod 2 & Blue disk on top of gray disk. Gray disk on top of green disk. Green disk in rod 1. The disks can be moved in rod 1, rod 2, rod 3. & 3.3 & 7.5 \\
\midrule
Ravens (Put~Blocks~in~Bowls) & Put the yellow blocks in gray bowls & There is a gray bowl 1, gray bowl 2, gray bowl 3, yellow block 1, yellow block 2, yellow block 3, blue bowl 1, red block 1, green bowl 1, orange block 1. & 6.1 & 25 \\
\midrule
BabyAI (Pickup) & Pick up the ball & Room 1 has purple ball. Room 2 has yellow key, agent. Room 3 has red key. The door connecting Room 1 and Room 2 is locked. The door connecting Room 2 and Room 3 is locked. & 6.7 & 7.7 \\
\midrule
VirtualHome & Read book &  & 5.9 & 150 \\
\bottomrule
    \end{tabularx}
  \caption{Table displays tasks from each environment, average plan length, and average action space size $|\mathcal{A}|$. For VirtualHome, we do not specify an initial observation since it is hard to describe a room environment. Here, the action space varies with episodes, depending for instance on the number of objects.}
  \label{tab:table_of_envs}
  \vspace{-0.3cm}
\end{table}

\begin{table}[t]
\small
\centering
\begin{tabular}{ccccccccc}
\hline
\textbf{Setup} & \textbf{Say Model} & {\textbf{Greedy-Token}} & \multicolumn{3}{c}{\textbf{Greedy-Action}} & \multicolumn{3}{c}{\textbf{Beam-Action}} \\
% \hline
 & & & Say & SayCan & SayCanPay & Say & SayCan & SayCanPay \\
\hline
\multirow{2}{*}{\parbox{3.2cm}{\centering Ravens \\ (tower of hanoi)}} 
& Vicuna & 45 & 48 & 48 & 50 & 54 & 68 & $\mathbf{70}$ \\  
& Flan-T5 & 30 & 30 & 39 & 42 & 38 & $\mathbf{50}$ & $\mathbf{50}$ \\
\hline
\multirow{2}{*}{\parbox{3.2cm}{\centering Ravens \\ (put blocks in bowls)}} 
% & Vicuna & 40 & 51 & 59 & 64 & 42 & 52 & 54 \\
& Vicuna & 30 & 51 & 52 & 54 & 52 & 52 & $\mathbf{56}$ \\
& Flan-T5 & 96 & 96 & 96 & 96 & $\mathbf{98}$ & $\mathbf{98}$ & $\mathbf{98}$ \\
\hline
\multirow{2}{*}{\parbox{3.2cm}{\centering BabyAI \\ (pickup)}} 
& Vicuna & 59 & 62 & 81 & 88 & 72 & $\mathbf{94}$ & $\mathbf{94}$ \\  
& Flan-T5 & 0 & 0 & 30 & $\mathbf{36}$ & 1 & $\mathbf{36}$ & 30 \\
\hline
\multirow{2}{*}{\parbox{3.2cm}{\centering VirtualHome}} 
& Vicuna & 0 & 32 & 49 & 52 & 48 & 52 & $\mathbf{53}$ \\  
& Flan-T5 & 0 & 0 & 30 & 48 & 30 & 41 & $\mathbf{50}$ \\
\hline
\end{tabular}
\caption{Table shows the \emph{planning success} (i.e. \# plans out of 100 that reached the goal within limited steps) on the test split across different environments using Vicuna, Flan-T5 models. It can be observed that the best decoding strategy is Beam-Action and the best decoding score is SayCanPay.}
\label{table-test-success}
% \vspace{-8pt}
\end{table}

%%%%%%%%%%%%%%%%%%%%%%%%%%%%%%%%%%%%
\begin{table}[t]
\small
\centering
\begin{tabular}{ccccccccc}
\hline
\textbf{Setup} & \textbf{Say Model} & {\textbf{Greedy-Token}} & \multicolumn{3}{c}{\textbf{Greedy-Action}} & \multicolumn{3}{c}{\textbf{Beam-Action}} \\
% \hline
 & & & Say & SayCan & SayCanPay & Say & SayCan & SayCanPay \\
\hline
\multirow{2}{*}{\parbox{3.2cm}{\centering Ravens \\ (tower of hanoi)}} 
& Vicuna & 12 & 24 & 55 & $\mathbf{58}$ & 20 & 47 & 52 \\  
& Flan-T5 & 34 & 34 & 46 & 47 & 38 & 54 & $\mathbf{56}$ \\
\hline
\multirow{2}{*}{\parbox{3.2cm}{\centering Ravens \\ (put blocks in bowls)}} 
& Vicuna & 16 & 36 & 40 & 48 & 38 & 42 & $\mathbf{56}$ \\  
& Flan-T5 & 63 & 65 & 71 & $\mathbf{74}$ & 67 & $\mathbf{74}$ & $\mathbf{74}$ \\
\hline
\multirow{2}{*}{\parbox{3.2cm}{\centering BabyAI \\ (pickup)}} 
& Vicuna & 48 & 50 & 53 & 54 & 56 & 56 & $\mathbf{62}$ \\  
& Flan-T5 & 0 & 0 & 26 & 28 & 1 & 30 & $\mathbf{34}$ \\
\hline
\multirow{2}{*}{\parbox{3.2cm}{\centering VirtualHome}} 
& Vicuna & 0 & 14 & 23 & 29 & 20 & 26 & $\mathbf{30}$ \\  
& Flan-T5 & 0 & 0 & 6 & 15 & 4 & 19 & $\mathbf{26}$ \\
\hline
\end{tabular}
\caption{Table shows the \emph{cost-effectiveness} (i.e. \#plans out of 100 that reached the goal within limited steps and also had the same plan length as the expert plan) on the \textbf{test} split across different environments using Vicuna, Flan-T5 models. It can be observed that the best decoding strategy is Beam-Action and the best decoding score is SayCanPay.}
\label{table-test-optimal}
\end{table}

%%%%%%%%%%%%%%%%%%%%%%%%%%%%%%%%%%%%%%
\begin{table}[t]
\small
\centering
\begin{tabular}{ccccccccc}
\hline
\textbf{Setup} & \textbf{Say Model} & {\textbf{Greedy-Token}} & \multicolumn{3}{c}{\textbf{Greedy-Action}} & \multicolumn{3}{c}{\textbf{Beam-Action}} \\    
% \hline
 & & & Say & SayCan & SayCanPay & Say & SayCan & SayCanPay \\
\hline
\multirow{2}{*}{\parbox{3.2cm}{\centering Ravens \\ (tower of hanoi)}} 
& Vicuna & 32 & 30 & 18 & 18 & 27 & $\mathbf{34}$ & $\mathbf{34}$ \\  
& Flan-T5 & 24 & 22 & 18 & 16 & $\mathbf{26}$ & $\mathbf{26}$ & $\mathbf{26}$ \\
\hline
\multirow{2}{*}{\parbox{3.2cm}{\centering Ravens \\ (put blocks in bowls)}} 
& Vicuna & 8 & $\mathbf{30}$ & 10 & 6 & $\mathbf{30}$ & 10 & 6 \\  
& Flan-T5 & 94 & 94 & 26 & 18 & $\mathbf{96}$ & 22 & 24 \\
\hline
\multirow{2}{*}{\parbox{3.2cm}{\centering BabyAI \\ (pickup)}} 
& Vicuna & 0 & 1 & 4 & $\mathbf{12}$ & 9 & $\mathbf{12}$ & 10 \\  
& Flan-T5 & 0 & 1 & $\mathbf{28}$ & $\mathbf{28}$ & 1 & 15 & $\mathbf{28}$ \\
\hline
\multirow{2}{*}{\parbox{3.2cm}{\centering VirtualHome}} 
& Vicuna & 0/20 & 2/20 & 3/20 & 3/20 & $\mathbf{5/20}$ & $\mathbf{5/20}$ & $\mathbf{5/20}$ \\  
& Flan-T5 & 0/20 & 0/20 & 0/20 & 3/20 & 1/20 & 3/20 & $\mathbf{5/20}$ \\
\hline
\end{tabular}
\caption{Table shows the generalization results (i.e. the number of plans out of 100 that reached the goal) on \emph{test-generalize} split across different environments using Vicuna and Flan-T5 models. It can be observed that Beam-Action outperforms other decoding strategies.}
\label{table-test-generalize}
\end{table}

%%%%%%%%%%%%%%%%%%%%%%%%%%%%%%%%%%%%%%%%%%%%%

\subsection{Pay Model}
\label{subsection:pay_model}
We model it as a regression problem to estimate action payoffs. Using expert trajectories $\mathcal{E}$, we create a dataset with each batch as $[g, h_{t-1}, a_t, r]^{1:B}$. Given sparse rewards (i.e. $r_{T}=1$), we use temporal discounting $\delta \in (0,1)$ to assign discounted rewards to previous actions in the trajectory\footnote{$\delta$ for the Pay model training is unrelated to the POMDP.}. This ensures that actions closer to the end receive higher rewards and vice versa. Specifically, $r_{T-1} = \delta, r_{T-2} = \delta^2$, and so on. We also sample negative actions from other paths (akin to the Can model) with a reward of $0$. The model is trained to align the discounted reward of the action and the predicted reward from $\mathcal{M}^{\text{pay}}$ by minimizing the mean squared error (MSE) loss $ \frac{1}{B} \sum_{i=1}^{B} (r^i - \mathcal{M}_{\text{pay}}(g^i, h_{t-1}^i, a_t^i))^2$. The model uses an uncased Bert plus a regression layer whose output is bounded in $[0, 1]$ via a sigmoid activation. The input format is the same as the Can model. The output is the estimated payoff, $f_{\text{heur}}(h_t, g) = \mathcal{M}^{pay}(g, h_{t-1}, a_t)$.\\

\noindent \textbf{Input} \code{$\langle$Goal$\rangle$} pick up the purple box. \code{$\langle$Initial State$\rangle$} Room 1 has yellow key, agent. Room 2 has purple box. The door connecting Room 1 and Room 2 is locked. \code{$\langle$Step 1$\rangle$} pick up yellow key. \code{$\langle$Step 2$\rangle$} toggle yellow door. \code{$\langle$Step 3$\rangle$} drop key in void. \code{$\langle$Step 4$\rangle$} pick up blue box. \code{$\langle$NXT$\rangle$} done picking up.\\
\textbf{Output} \answer{1.0} \quad\quad// \emph{end of plan}

\noindent \textbf{Input} \code{$\langle$Goal$\rangle$} pick up the purple box. \code{$\langle$Initial State$\rangle$} Room 1 has yellow key, agent. Room 2 has purple box. The door connecting Room 1 and Room 2 is locked. \code{$\langle$Step 1$\rangle$} pick up yellow key. \code{$\langle$Step 2$\rangle$} toggle yellow door. \code{$\langle$Step 3$\rangle$} drop key in void. \code{$\langle$NXT$\rangle$} pick up blue box.\\
\textbf{Output} \answer{0.6} \quad\quad$// \delta\cdot r$  

\noindent \textbf{Input} \code{$\langle$Goal$\rangle$} pick up the purple box. \code{$\langle$Initial State$\rangle$} Room 1 has yellow key, agent. Room 2 has purple box. The door connecting Room 1 and Room 2 is locked. \code{$\langle$Step 1$\rangle$} pick up yellow key. \code{$\langle$Step 2$\rangle$} toggle yellow door. \code{$\langle$Step 3$\rangle$} drop key in void. \code{$\langle$NXT$\rangle$} pick up green box.\\
\textbf{Output} \answer{0} \quad\quad// \emph{very low payoff}\\
%%%%%%%%%%%%%%%%%%%%%%%%%%%%%%%%%%%%%%%%%%%%%%%%%

\begin{figure*}[t]
    \centering
    \includegraphics[width=\linewidth]{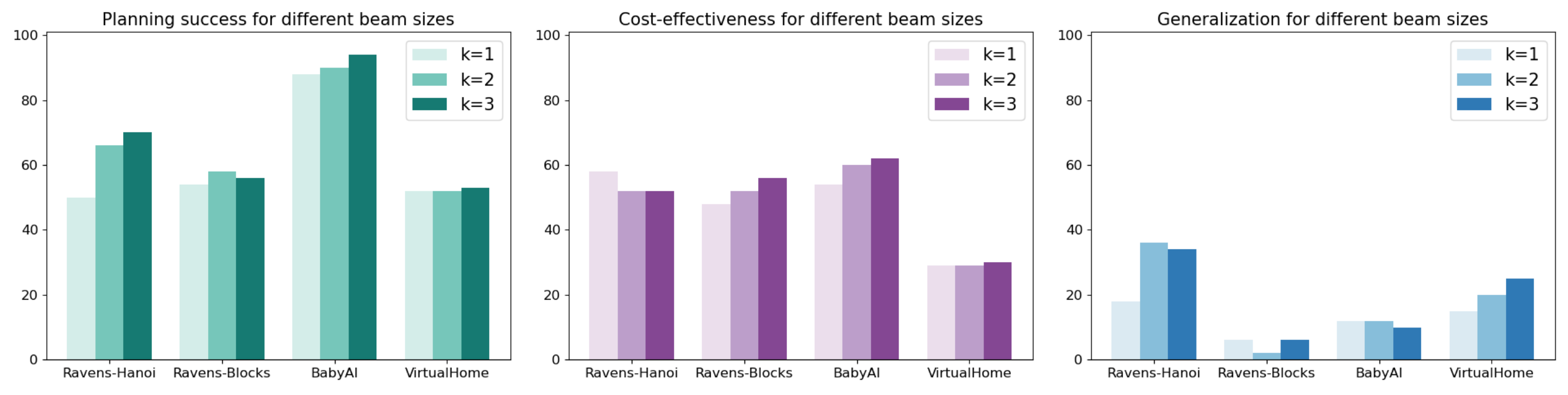}
    \caption{[Best viewed in color] From left to right: Planning success, cost-effectiveness, generalization for different beam sizes. Note, that generalization on the test-generalize split for VirtualHome is reported as a percentage.}
    \label{fig:beam-comparison}
\end{figure*}

%%%%%%%%%%%%%%%%%%%%%%%%%%%%%%%%%%%%%%%%%%%%%%%%%%%%%%%%%%%%%
\section{Experimental Setup}

\subsection{Say Model}
\label{subsection: say_model}
The Say model does not undergo any fine-tuning and is used only for inference. We experimented with two types of transformer architectures. (i) \textbf{Decoder type}: 13b-parameter Vicuna model~\cite{vicuna2023} trained by fine-tuning LLaMA~\cite{llama2023}. (ii) \textbf{Encoder-decoder type}: Flan-T5-11b~\cite{flan-t5} which is the instruction fine-tuned version of the T5 transformer~\cite{t5-transformer}. Existing works have demonstrated the planning abilities of both the decoder type~\cite{plansformer} and the encoder-decoder type architectures~\cite{valmeekam2023planning,llms_cant_plan}.
Since the generated plan is in free-form language and may contain unrecognizable (for the environment) words or incorrect syntax, it cannot be directly translated into actionable steps in the environment. Following \citet{llms_zero_shot}, we use an exhaustive list of admissible actions (feasible and otherwise), and at the end of each action step, map the generated action to the closest admissible action using minimum edit distance. Interleaving action generation and mapping ensures that all subsequent steps are conditioned on admissible actions, thus mitigating compounding errors. We provide 3 examples (input goal and observation, output plan) to the model via few-shot prompting. 

\subsection{Environments}
\label{subsection:environments}
% prompting
% show a table of input + prompts?
We tested in three environments, detailed in Table~\ref{tab:table_of_envs}.

\begin{itemize}[leftmargin=*,noitemsep]
    \item \textbf{Ravens}~\cite{ravens} is a PyBullet simulated task set focusing on ``pick and place". It includes 10 tabletop tasks, of which we use two: (i) Tower of Hanoi (sequence), a variation of the classic puzzle focusing on specific intermediate goals, like moving a particular disk to a designated rod while keeping the traditional constraints. This creates more goal diversity; (ii) Put blocks in bowls requires placing blocks into bowls based on rules like \emph{put yellow block in green bowls}. We adapt the environment for language tasks, observations, and actions.
    \item \textbf{BabyAI}~\cite{babyai_iclr19} is a 2D-gridworld environment where a bot is provided a language task sampled from a predefined grammar. We focus on \emph{pickup} tasks where the agent navigates to collect an object, often unlocking doors or moving obstacles. Task difficulty varies with rooms, obstacles, and distractor objects. The agent's actions include high-level commands like \emph{pickup} and \emph{drop} which are composed of atomic actions: ``left", ``right", ``forward", ``pick", and ``drop" (see Figure~\ref{fig:saycanpay_teaser})
    \item \textbf{VirtualHome}~\cite{virtualhome} is an interactive platform to simulate complex household activities via interactions with the environment, such as picking up objects, switching on/off appliances. We utilize the VirtualHome-Env dataset~\cite{virtualhomeDataset}, comprising daily household activities from 7 scenes gathered via crowdsourcing. We only use the goal as the input (see Table~\ref{tab:table_of_envs}).
\end{itemize}

\paragraph{Data Splits and Evaluation.} We aim to assess the success, cost-effectiveness, and out-of-distribution (OOD) generalization of the generated plans. We created three data splits for each environment using expert trajectories. (i) \textbf{train} split for Can, Pay model training and few-shot prompting of the Say Model; (ii) \textbf{test} split assesses the LM planners' ability to generate successful plans (i.e. reach the goal within limited steps), and also the planners' ability to generate cost-effective plans (i.e. plans that succeed and also have the same plan length as the expert plan\footnote{We split test into two parts of 100 samples to evaluate success, cost-effectiveness. For VirtualHome, we use the annotated plans from its dataset.}). (iii) \textbf{test-generalize} split focuses on the generalization capabilities like handling novel initial observations (e.g., unseen colors of blocks and bowls, distractors in BabyAI), longer sequence lengths (e.g., more blocks or disks in Ravens, more rooms in BabyAI), and unseen tasks in VirtualHome. All test splits have \# total episodes = 100 unless specified otherwise. Moreover, all splits are disjoint (i.e. no overlap).

\paragraph{Baselines.} At the action level, we evaluate our decoding scores (Say, SayCan, SayCanPay) using various decoding strategies (Greedy and Beam-Action). Therefore, our baselines employ a mix of these strategies and scores. For tokens, we use the Greedy-Token decoding strategy as a reference. Notably, Greedy-Action SayCan is the offline planning version of the original SayCan paper~\cite{saycan}.

\paragraph{Training and Inference Details.} 
We use 800 expert train trajectories for each Ravens task and 400 for BabyAI. For VirtualHome, we retained $\approx 800$ compatible trajectories for the current simulator. An additional 100 expert trajectories were collected for each test split (20 for VirtualHome test-generalize). The Can and Pay models were trained on 7 NVIDIA-DGX V-100 GPUs using the Distributed Data-Parallel framework across 20 epochs. Training parameters included a 1e-4 learning rate, AdamW optimizer with 1e-5 weight decay, a batch size of 50, a train-validation split of 80-20. For inference, the Say model was loaded using Model Parallel on the same GPUs. Inference hyperparameters are listed in Table~\ref{tab:inference hyperparameters}. Parameters like beam groups and diversity penalty encourage diversity among the beams, thus avoiding multiple similar sequences. We used 8-bit precision for GPU-efficient model loading~\cite{llm-8bit}.

\subsection{Results}
\label{subsection:results}
% Qualitative analysis of probabilities
We analyze the results along the following axes: decoding strategies, decoding scores, and transformer architectures. We assessed planning success and generalization by executing the generated plans in simulators such as Ravens and BabyAI, which have built-in validation checks to determine goal achievement. For the more open-ended VirtualHome environment, we manually reviewed fully executed plans to ensure they met the intended task objectives. For cost-effectiveness, we acquired expert trajectories for each test sample using an oracle planner.

\noindent\textbf{Comparing decoding scores.} From Tables~\ref{table-test-success},~\ref{table-test-optimal}, the performance across various decoding scores can be summarized as Say~$<$~SayCan~$\leq$~SayCanPay. \textbf{(i)~planning success}: The SayCanPay and SayCan scores lead to comparable performances, often outperforming Say. The Pay model's minor performance edge could be due to its focus on selecting actions based on long-term relevance, potentially avoiding irreversible (\emph{breaking an egg}) or even absorbing states (\emph{discharged cellphone}) from where it is impossible to reach the goal (i.e. planning is non-ergodic). \textbf{(ii) cost-effectiveness}: SayCanPay leads to a significant improvement over both Say ($\approx 11-97 \%$ for Beam-Action) and SayCan ($\approx 0-33 \%$ for Beam-Action and $\approx 1-150 \%$ for Greedy-Action). \textbf{(iii) generalization:} From Table~\ref{table-test-generalize}, while the overall performance for SayCan and SayCanPay improves over Say, a noticeable drop in performance was observed for Ravens. This led to the hypothesis that the learned domain models (Can, Pay) are not generalizing to OOD data in certain environments (see \S~\ref{subsection:limitations_and_future_work} for potential solutions). 
% \textbf{(iv) relative length:} Figure~\ref{fig:relative length} demonstrates that the SayCanPay scoring achieves the best relative length ($\approx 1$) for both Greedy and Beam-Action strategies signifying the cost-efficiency of the generated plans.

\paragraph{Comparing decoding strategies.}
From Tables~\ref{table-test-success},~\ref{table-test-optimal},~\ref{table-test-generalize}, the overall performance across decoding strategies follows the pattern: Greedy-Token $<$ Greedy-Action $<$ Beam-Action across all splits. The Beam-Action Say, SayCan, and SayCanPay versions show improvement over their corresponding Greedy-Action counterparts. \textbf{(i) planning success:} Beam-Action SayCanPay beats Greedy-Action SayCanPay by $\approx~1-40 \%$. Similar gains are also observed with other decoding scores. \textbf{(ii) cost-effectiveness:} Beam-Action SayCanPay improves over Greedy-Action SayCanPay by $\approx~0-73 \%$. \textbf{(iii) generalization:} Beam-Action SayCanPay beats Greedy-Action SayCanPay by $\approx~0-89 \%$. 
% \textbf{(iv) relative length:} As shown in Figure~\ref{fig:relative length}, Beam-Action shows a slight improvement over Greedy-Action. This demonstrates the importance of heuristic search in achieving success, cost-efficiency, and generalization.

\paragraph{Comparing Transformer Architectures.} We did not observe a consistent performance gain for any particular architecture, suggesting that either is apt for planning. We lack a definitive explanation, and further research is required to understand how each LM component impacts reasoning.

\begin{figure*}[t]
    \centering
    \includegraphics[width=0.9\linewidth]{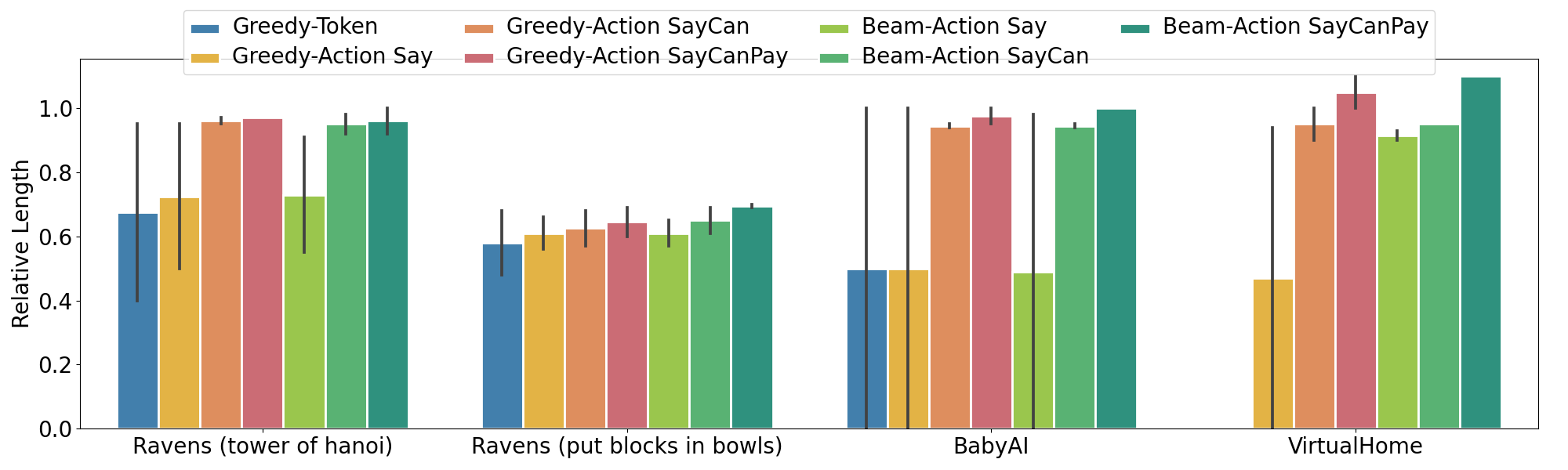}
    \caption{[Best viewed in color] The error plot represents the variance in relative length over models Vicuna and Flan-T5. Due to the open-ended nature of VirtualHome, the crowdsourced trajectories are not optimal, which explains why certain cases have a relative length $>1.0$. Note that Greedy-Token decoding in VirtualHome has a relative length $=0$ since no generated plans were executed successfully for both Vicuna and Flan-T5.}
    \label{fig:relative length}
\end{figure*}

\subsection{Ablation Details}

\begin{itemize}
    \item Effect of beam-size $k$: As seen in Figure~\ref{fig:beam-comparison}, in general, both plan success and cost-effectiveness increases with increase in beam size with $k=$ 1 (Greedy-Action), 2, 3 (Beam-Action). All experiments used the SayCanPay decoding score. However, no clear patterns were observed for generalization results.

    \item Impact of Say Model: Planning failures may arise because the Say model fails to propose a right action amongst the candidates, making Can and Pay ineffective. We studied the Say model's impact on overall performance using a \emph{Perfect Say} that always recommends the correct action along with random distractors. From Table~\ref{tab:perfect-say}, we observed $16$-$84\%$ improvements in SayCan and SayCanPay performance across various environments, indicating the potential of an improved Say model. Thus, using a larger model trained on more data could potentially enhance performance.

    \item Plan length comparison: We compute a \emph{relative length}$=$ \emph{oracle plan length / generated plan length}, which compares the generated and oracle plan lengths. A value $=1$ indicates equal lengths and a value $=0$ that the plan length is infinity (i.e. an unsuccessful plan). As shown in Figure~\ref{fig:relative length}, Beam-Action slightly improves over Greedy-Action. Furthermore, SayCanPay scoring achieves the best relative length ($\approx 1$) for both Greedy and Beam-Action strategies signifying the cost-efficiency of the generated plans.

    \item Impact of problem size on planning time. Effect of action space: Planning time remains unaffected since the Say model generates rather than discriminates between actions. Effect of plan length: Greedy-Token run time increases by \(\sim \)2s for each action step. Effect of decoding strategy: \(\sim \)9s for Greedy-Token, \(\sim \)17s for Greedy-Action, \(\sim \)35s for Beam-Action. Effect of decoding score: Planning time is unaffected since the Can and Pay are small LMs with negligible overheads. Quantization techniques and advanced hardware can further reduce run time and is an active research area~\cite{qlora,gptq}.

    \item Qualitative Analysis: The Can model effectively selects feasible actions (Figure~\ref{fig:saycanpay_teaser}). The Pay model prioritizes actions that lead to quicker goal achievement. While Pay gives a high probability to the ``done task" action linking it to plan completion, the Can score negates it due to unsatisfied ``done task" preconditions.
\end{itemize}

%%%%%%%%%%%%%%%%%
% \begin{wraptable}{r}{0.52\linewidth}
    \begin{table}[ht]
    \centering
    \begin{tabular}{lcl}
    \hline
    \textbf{Parameter} & \textbf{Value} & \textbf{Exceptions}\\
    \hline
    \multirow{2}{*}{max new tokens} & \multirow{2}{*}{10} & 11 Vicuna (Ravens-Blocks), \\
    & & 3 (VirtualHome)\\
    beam groups & 3 & 4 for Flan-T5 (BabyAI) \\
    diversity penalty & 2.0 & \\
    \hline
    candidates ($m$) & 6 & 8 for Flan-T5 (Baby-AI) \\
    beam-size ($k$) & 3 & \\
    \hline
    \end{tabular}
    \caption{Inference hyperparameters. Here the candidates ($m$) and the beam-size ($k$) parameter are over actions. The rest of the beam search parameters are over tokens.}
    \label{tab:inference hyperparameters}
    \end{table}
% \end{wraptable}

%%%%%%%%%%%%%%%%%%%%%%%%%%%%%%%%%%%%%%%

\begin{wraptable}{r}{0.5\linewidth}
% \begin{table}[t]
\begin{tabular}{llcc}
 & \textbf{Score} & \textbf{LM} & \textbf{Perfect} \\
 \hline
\multirow{2}{*}{Ravens-Hanoi} & SayCan & 48 & 88 \\
 & SayCanPay & 50 & 92 \\
 \hline
\multirow{2}{*}{Ravens-Blocks} & SayCan & 52 & 70 \\
 & SayCanPay & 54 & 75 \\
 \hline
\multirow{2}{*}{BabyAI} & SayCan & 81 & 90 \\
 & SayCanPay & 88 & 92 \\
 \hline
\multirow{2}{*}{VirtualHome} & SayCan & 49 & 60 \\
 & SayCanPay & 52 & 64\\
\hline
\end{tabular}
\caption{The table depicts the impact of the Say model on planning success performance. In this context, both ``LM" and ``Perfect" represent Say models. ``LM" corresponds to the Vicuna model, while ``Perfect Say" is an oracle Say model that consistently proposes the correct action along with two other distractor actions as next candidates. For all experiments, we used the Greedy-Action decoding strategy.}
\label{tab:perfect-say}
% \end{table}
\end{wraptable}

%%%%%%%%%%%%%%%%%%%%%%%%%%%%%%%%%%%%%%%%%%%%%%%

\subsection{Limitations and Future Work}
\label{subsection:limitations_and_future_work}
The main limitations are (i) the need for expert trajectories to train domain models, and (ii) the domain models' limited adaptability to OOD data. These challenges are inherent to deep learning models. However, recent advances in LLMs offer promising solutions. For example, newer studies have leveraged LLMs for reward design due to their ability to infer intentions from minimal prompts~\cite{LLM_reward_design}. Notably, LLMs outperform smaller counterparts like Bert in generalization. Since both Can and Pay scores resemble rewards, future studies could use LLMs to mitigate training and improve generalization. Another potential direction could be to experiment with symbolic methods and non-parameterized heuristics like comparing the current generated plan with the successful/expert trajectories in the buffer.

%%%%%%%%%%%%%%%%%%%%%%%%%%%%%%%%%%%%%%%%%%%%%%%%%%%
\section{Conclusion}
\label{subsection:conclusion}
We proposed to combine the world knowledge and generative capabilities of LLMs with the systematicity of classical planning by formulating a heuristic search-based planning framework for LLMs. We demonstrated how to generate plans that are both feasible and cost-effective. While LLMs still cannot generate long-horizon plans on par with classical planners, our method overcomes issues inherent to LLM-based planning and extends traditional planning with the advantages of language models, marking significant progress for planning research with LLMs.

\section*{Acknowledgement}
This work was supported by the Wallenberg AI, Autonomous Systems and Software Program (WASP) funded by the Knut and Alice Wallenberg Foundation, and is also part of the EU H2020 ICT48 project “TAILOR” under contract 952215, and the KU Leuven Research Fund (C14/18/062).

\bibliography{main}

\end{document}